\documentclass{article}
\usepackage{amsmath,epsfig}
\usepackage[preprint]{spconf}

\copyrightnotice{978-1-6654-3864-3/21/\$31.00 ©2021 IEEE}

\let\OLDthebibliography\thebibliography
\renewcommand\thebibliography[1]{
  \OLDthebibliography{#1}
  \setlength{\parskip}{0pt}
  \setlength{\itemsep}{0pt plus 0.3ex}
}

\PassOptionsToPackage{hyphens}{url} 

\usepackage{times}
\usepackage{epsfig}
\usepackage{graphicx}
\usepackage{amsmath}
\usepackage{amssymb}

\usepackage{etoolbox}
\newtoggle{isDraft}

\usepackage{comment}
\usepackage{color}
\usepackage{csquotes}
\usepackage{multirow}
\usepackage{footnote}
\usepackage{rotating}
\usepackage{mwe,tikz}
\usepackage[percent]{overpic}

\usepackage{cite}

\usepackage{hyperref}
\usepackage{xcolor}
\hypersetup{
    colorlinks,
    citecolor={green!75!black},
    linkcolor={red!75!black}
}

\usepackage{cleveref}
\usepackage{float}
\usepackage{ntheorem}

\crefname{figure}{Figure}{Figures}
\crefname{table}{Table}{Tables}
\crefname{section}{Section}{Sections}

\newcommand{\etal}{\textit{et al.}}

\newcommand{\mesonet}{\emph{MesoInc-4}}
\newcommand{\simplelinear}{\emph{Durall \etal}}
\newcommand{\easyspot}{\emph{Wang \etal}}
\newcommand{\misl}{\emph{Bayar \etal}}
\DeclareMathOperator{\C}{C}
\let\S\relax

\let\B\relax
\DeclareMathOperator{\S}{S}

\DeclareMathOperator{\B}{B}
\DeclareMathOperator{\CS}{CS}
\DeclareMathOperator{\PP}{PP}
\DeclareMathOperator{\ct}{ct}
\DeclareMathOperator{\cts}{cts}
\DeclareMathOperator{\CST}{CST}
\newcommand{\setff}{FaceForensics++}

\newcommand \DSName {{VideoForensicsHQ}}
\newcommand{\setdf}{DF}
\newcommand{\setftf}{F2F}
\newcommand{\setfs}{FS}
\newcommand{\setnt}{NT}
\newcommand{\setdeeper}{DeeperForensics 1.0}

\DeclareMathOperator{\In}{In}
\DeclareMathOperator{\Out}{Out}
\DeclareMathOperator{\M}{M}

\DeclareMathOperator{\thr}{thr}

\makeatletter
\newcommand{\pushleft}[1]{\ifmeasuring@#1\else\omit$\displaystyle#1$\hfill\fi\ignorespaces}
\makeatother

\usepackage[most]{tcolorbox}
\definecolor{stableYellow}{rgb}{1.0, 1.0, 0.67}
\definecolor{stableGreen}{rgb}{0.85, 1.0, 0.85}
\definecolor{stableRed}{rgb}{1.0, 0.85, 0.85}
\newtcolorbox{stable}{colback=stableYellow,grow to right by=0mm,grow to left by=0mm, boxrule=0pt,boxsep=0pt,breakable}\definecolor{stableGreen}{rgb}{0.85, 1.0, 0.85}
\newtcolorbox{done}{arc=0pt,outer arc=0pt,boxsep=0pt,frame hidden,left=0pt,right=0pt,top=0pt,bottom=0pt,colback=stableGreen,colframe=stableGreen,grow to right by=0mm,grow to left by=0mm, boxrule=0pt,boxsep=0pt,breakable}
\newtcolorbox{unstable}{colback=stableRed,grow to right by=0mm,grow to left by=0mm, boxrule=0pt,boxsep=0pt,breakable}

\newcommand{\myparagraph}[1]{\vspace{1.5pt}\noindent{\textbf{#1}}}
 
\newcommand{\note}[1]{\textbf{Note}: #1}

\theoremseparator{:}
\newtheorem*{hypo_main_thm}{Hypothesis (H)}
\newenvironment{hypo_main}{\begin{hypo_main_thm}\label{hyp:main}}{\end{hypo_main_thm}}
\newcommand{\refhm}{{\hypersetup{hidelinks}{\hyperref[hyp:main]{\textbf{H}}}}}

\newcommand{\arch}{Arch.}

\newcommand{\acc}[1]{Acc.#1}

\newcommand{\setSA}{Group$\#1$}
\newcommand{\setRD}{Group$\#2$}
\newcommand{\setYT}{Group$\#3$}

\DeclareMathOperator{\sigmoid}{sigmoid}

\begin{document}\topmargin=0mm\sloppy

\title{\uppercase{VideoForensicsHQ: Detecting High-quality Manipulated Face Videos}}
\name{ \parbox{\linewidth}{\centering Gereon Fox, Wentao Liu, Hyeongwoo Kim, Hans-Peter Seidel, \\ Mohamed Elgharib \& Christian Theobalt}}
\address{Max Planck Institute for Informatics, Saarland Informatics Campus \\ \small\texttt{\{gfox,wliu,hyeongwoo.kim,hpseidel,elgharib,theobalt\}@mpi-inf.mpg.de}}
\maketitle

\begin{abstract}
There are concerns that new approaches to the synthesis of high quality face videos may be misused to manipulate videos with malicious intent. The research community therefore developed methods for the detection of modified footage and assembled benchmark datasets for this task.
In this paper, we examine how the performance of forgery detectors depends on the presence of artefacts that the human eye can see. We introduce a new benchmark dataset for face video forgery detection, of unprecedented quality. It allows us to demonstrate that existing detection techniques have difficulties detecting fakes that reliably fool the human eye. We thus introduce a new family of detectors that examine combinations of spatial and temporal features and outperform existing approaches both in terms of detection accuracy and generalization.

\end{abstract}

\section{Introduction}
\label{sec:intro}

Model- and learning-based approaches to face video 
 synthesis have reached high levels of visual realism. 
 Some  allow facial expressions to be modified or transferred \cite{thies2016face,kim2018DeepVideo,thies2019}, while others implement face swapping, i.e. replacing the face interior with a different face identity~\cite{Garrido2014}. 
 Reacting to concerns that these could be misused 
to modify videos in unethical ways, 
 the research community has developed techniques to detect forgeries,
 for generic content \cite{Bayar18,Afchar2018,wang2019cnngenerated} as well as specifically for faces \cite{roessler2019faceforensics++,Agarwal19,Ekraam19%,durall2019unmasking
 }.
In order to compare the effectiveness of forgery detection methods it is vital to evaluate them on benchmark datasets. 
As one example,  FaceForensics++~\cite{roessler2019faceforensics++} contains internet videos modified by several face synthesis techniques~\cite{DeepFakes,Faceswap,thies2016face,thies2019,Google} and demonstrates that an off-the-shelf image classifier, XceptionNet \cite{Chollet17}, outperforms methods specifically designed for fake detection.
However, whenever a forgery detector achieves a high detection accuracy on a dataset, we must wonder: Does this mean that the detector is very good,
or does it mean that the fakes in the dataset are just too easy
to detect? Based on the observation that the fakes in
existing benchmark datasets of forged face videos seem to be easy to spot for 
the \emph{human} eye (\cref{fig:datasets}), we have formulated the following hypothesis:
\begin{figure}%[H]
        \centering
         \includegraphics[width=0.95\linewidth]{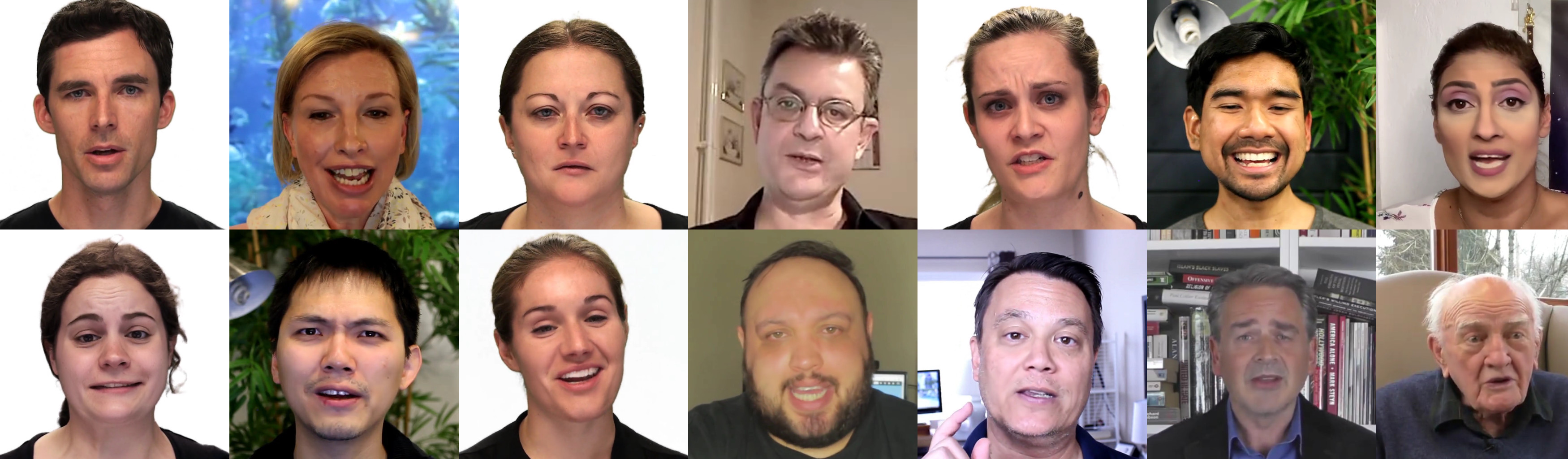}
        \caption{Forgeries from \DSName.
        }
        \label{fig:images}
        
\end{figure}
\begin{hypo_main}
\label{hyp:one}
The accuracy of existing face video forgery detection methods depends on   visual artefacts that humans would be able to spot with the naked eye. As soon as fakes are missing such artefacts, detector performance will drop.
\end{hypo_main}
The artefacts in question include temporal jitter, implausible lighting, unnatural smoothness and blending boundaries, occuring as part of the synthesis process.
In the course of investigating \refhm{} we make two main contributions:

First, we present \DSName{}, a benchmark dataset of high quality face video 
 manipulations, designed to \emph{not} include said artefacts (\cref{fig:images}). Our user study shows (see supplemental) that humans find our fakes considerably harder to detect than in previous datasets. Only \DSName{} allows us to investigate \refhm{}, by evaluating existing detectors on it, showing that their performance leaves room for improvement.
Second, making use of this room, we present a novel family of learning-based detectors that examine combinations of color, low-level noise and temporal correlations. We find these to perform better than previous methods on high-quality fakes and to even generalize to unseen synthesis methods.

\section{Related Work}
\label{sec:related}

\begin{figure}
\centering
\begin{tabular}{cc}
\includegraphics[width=.45\linewidth]{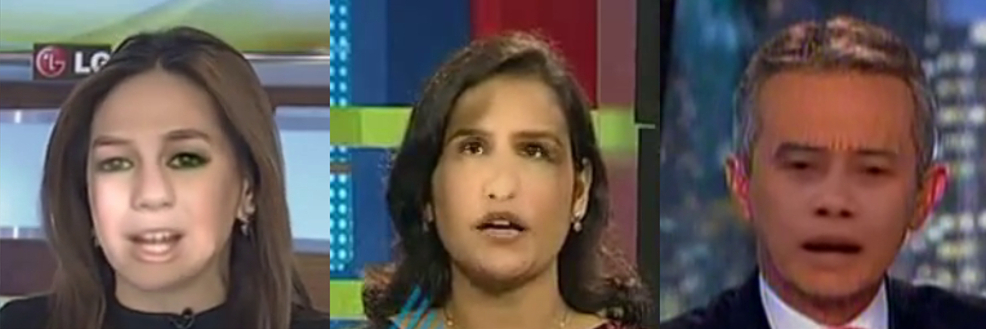} & \includegraphics[width=.45\linewidth]{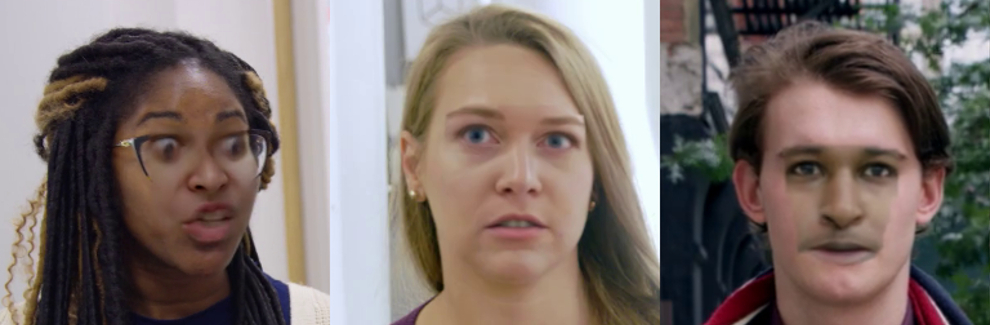}  \\
\setfs, \setdf, \setftf \cite{roessler2019faceforensics++} &  Deep Fake Dataset\cite{Google} \\
\includegraphics[width=.45\linewidth]{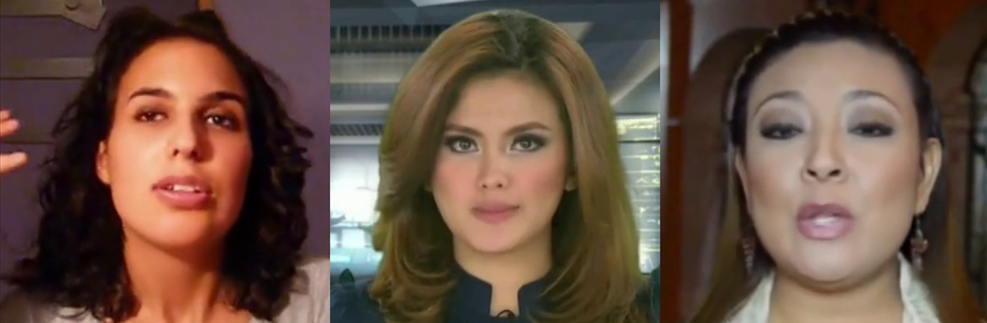} & \includegraphics[width=.45\linewidth]{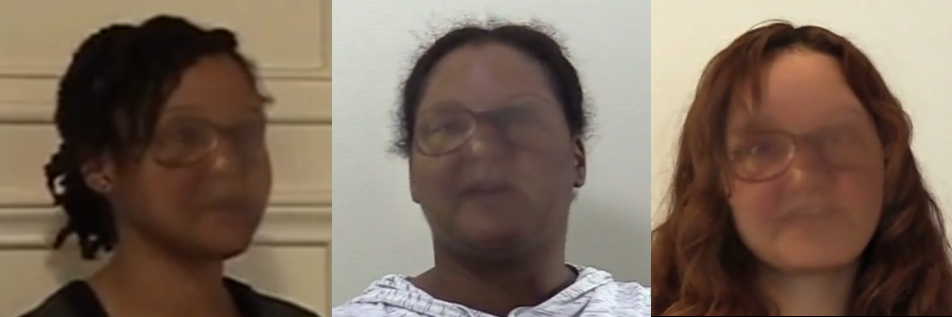}  \\
\setnt \cite{roessler2019faceforensics++} & DFDC \cite{dolhansky2019} \\
\end{tabular}
	\caption{Previous face video manipulation datasets contain noticeable artefacts. \enquote{Neural Textures} \cite{roessler2019faceforensics++} offers the best quality so far, which is why we included it in our user study.}
	\label{fig:datasets}
\end{figure}

\myparagraph{Face Reenactment \& Editing:} Face reenactment techniques control facial expressions in a video 
\cite{thies2016face,kim2018DeepVideo,Kim19NeuralDubbing, %,Nagano2018
%,Wiles18,
%Wu2018
Zakharov20
}. 
Many of them extract expressions by fitting a face model\cite{garrido2016reconstruction}
and then re-synthesizing the face with parameters copied from a source video \cite{thies2016face,kim2018DeepVideo,Kim19NeuralDubbing%,Nagano2018
}. 
Kim~\etal~\cite{kim2018DeepVideo} for the first time showed space-time coherent realistic global pose and expression editing in videos using a GAN. 
 \cite{Kim19NeuralDubbing} is a more comprehensive overview.

\myparagraph{Face Manipulation Datasets:} Several datasets of manipulated images \cite{Zhou17,Guan19} or videos~\cite{%Korshunov18,
Guan19,roessler2019faceforensics++,dolhansky2019} exist. 
 Roessler~\etal's FaceForensics++ dataset \cite{roessler2019faceforensics++} contains 1,000 videos, each manipulated by 4 different techniques \cite{DeepFakes,Faceswap,thies2016face,thies2019}.
  Their results show that many manipulation approaches
   produce very noticeable artefacts. 
Google released the Deep Fake Detection Dataset~\cite{Google}.
 Many sequences exhibit visual artefacts. %; audio is not provided.
 Facebook  released a dataset \cite{dolhansky2020} of manipulated face videos of varying quality, with face resolution often much less than $299^2$. 
 Jiang \etal{} presented \setdeeper{}~\cite{jiang2020}, which, augmentations aside, provides 1000 forgeries derived from \cite{roessler2019faceforensics++}.
As can be seen in \cref{fig:datasets}, our user study, and our evaluation (\cref{sec:results}), all these datasets are made up of fakes that are easily detected by humans and machines.

\myparagraph{Detection of Manipulated Visual Content:}
Generic detection techniques \cite{Fridrich12%,Cozzolino14
,Cozzolino17,%,Bayar16,%Rahmouni17,Li19,
Zhou18,Bayar18,Wu19,Cozzolino20,wang2019cnngenerated} often examine low-level features such as high-frequency components and noise.
Fridrich~\etal~\cite{Fridrich12} introduced convolutional kernels designed for steganalysis, that were later formulated as a constrained CNN \cite{Cozzolino17}.
Bayar~\etal~\cite{%Bayar16,
Bayar18} suppress image content to focus on low-level patterns. 
The work of \cite{Cozzolino20} localizes edited regions of an image by examining so-called \enquote{noiseprints}.
Face-specific techniques  can be classified into single-image-based \cite{Zhou17,Afchar2018,%Raghavendra17,
li2019_6,Agarwal19,roessler2019faceforensics++,%wang2019detecting,
li2019exposing,durall2019unmasking,wang2019cnngenerated} and multi-image-based~\cite{Ekraam19,du2019generalizable%,Nguyen19
} approaches. 
Zhou~\etal~\cite{Zhou17} detect face swaps using a two stream network (visual artefacts + steganalysis features).
\mesonet{} \cite{Afchar2018} is 
 a CNN 
 that learns at which level of granularity to investigate the input.
R\"ossler~\etal~\cite{roessler2019faceforensics++} examined a variety of manipulation detection techniques \cite{Fridrich12%,Cozzolino14
,Cozzolino17,%Bayar16,%Rahmouni17,
Afchar2018} on their FaceForensics++ dataset, 
 with XceptionNet~\cite{Chollet17} emerging as the most robust detector. 
Other works have studied temporal correlations, based on \enquote{action-units} \cite{Agarwal19} or RNN's \cite{Ekraam19}.

\section{The \DSName{} Dataset}
\label{sec:dataset}

\begin{figure*}
\centering
	\includegraphics[width=1.0\linewidth]{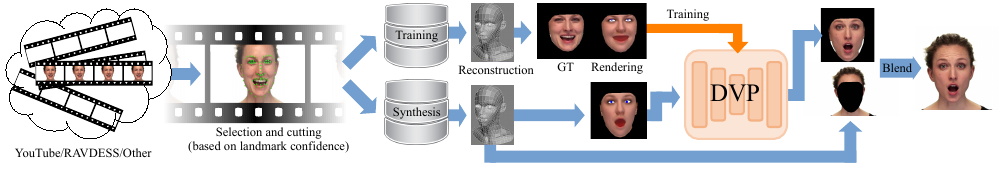}
	\vspace{-0.5cm}
	\caption{We apply monocular reconstruction to videos from the \enquote{Synthesis} set, to obtain facial parameters that are close to the training distribution. DVP \cite{kim2018DeepVideo}  turns these into photorealistic videos.
	}
	\label{fig:inference}
\end{figure*}

To investigate \refhm{}, we need a benchmark dataset that
contains \emph{many} fakes of high quality: In order for humans to be unable to spot fakes, we must avoid artefacts such as temporal jitter, unnatural movement, implausible lighting, unusual smoothness, or strong blending boundaries. While there definitely are state-of-the-art synthesis techniques that achieve such quality under ideal conditions, we are not aware of a \emph{large-scale} benchmark dataset that aggregates \emph{many} such high quality results.
\DSName{} is the first such dataset, as confirmed by our user study (supplemental material).
The challenge does \emph{not} lie in finding a novel synthesis method and it is by no means our goal to present one. Instead we adapt Deep Video Portraits (DVP) \cite{kim2018DeepVideo} for \emph{large-scale} fake creation. 

While DVP can transfer performances from a source person to a \emph{different} target person, this mode can  lead to artefacts if the distribution of facial expressions differs a lot between source and target. Not even the \enquote{style-preserving} variant \cite{Kim19NeuralDubbing} avoids glitches as reliably as necessary. 
  We thus produce \enquote{intra-person} transfers (i.e. source and target are the same person). Recent works \cite{Suwajanakorn2017,Fried2019} show this to be a very relevant threat-scenario.
   Since DVP is trained on a set of frames that is disjoint from the source/target sequence, it has not seen the expressions to synthesize in advance and must still generate the typical GAN artefacts that are common with synthesis methods, but typically go unnoticed by humans.

\myparagraph{\DSName~ At-A-Glance:}\label{sec:glance}
\DSName~contains 1737 videos of
talking faces ($43\%$ male, $57\%$ female), with 8 different emotions. Most videos have resolution $1280\times 720$. They amount to 1,666,816 frames with average resolution $968^2$ and the average face covering $487^2$ pixels.
There are three different subsets: \setSA{} was mined from~\cite{Kim19NeuralDubbing}, \setRD{} from RAVDESS~\cite{Livingstone2015TheRA}, and \setYT{} from YouTube. 
 In total, our dataset contains 326,973 fake frames, comparable to the \enquote{Neural Textures} \cite{thies2019} part of FaceForesnics++.
While their fakes are the ones that come closest to our dataset in terms of visual quality (see \cref{fig:datasets}), our user study (supplemental) shows that our fakes are much harder to detect for humans:  $65.8\%$ of our fakes are mistaken as reals, while only $14.3\%$ of the \enquote{Neural Textures} fakes pass this test.
For more details see our supplemental document.

\myparagraph{Production process:}\label{sec:creation}
Mining real videos as the basis for our fakes is challenging 
 because jump-cuts, animations and unusual face poses need to be circumvented automatically, especially for YouTube.
To synthesize video of an identity, DVP requires about 5 - 10 minutes of training material, with all frames showing the same face at roughly the same distance, in a near-frontal pose.
To find such material, we run a facial landmark tracker \cite{Saragih2011} on all frames of all source videos, obtaining 66 landmark positions for every frame $f_i$, and one confidence value in the range $[0, 1]$ for every landmark position. We compute three metrics: 
\begin{enumerate}
    \item $c_i$: avg. landmark confidence for frame $f_i$
    \item $d_i$: avg. offset between landmark positions in $f_i$ and   $f_{i - 1}$, divided by face size
    \item mean and standard deviation of the $c_i$'s and $d_i$'s
\end{enumerate}
A frame is regarded unsuitable in any of the following cases: a) $c_i < 0.2$, b) $d_i > 0.1$, c) $c_i < 0.6$ deviates from the confidence mean by more than $110\%$ of the std.dev. (in neg. direction), d) $d_i > 0.025$ deviates from the displacement mean by more than $110\%$ of the std.dev. (in pos. direction). If none of these apply, the frame is added to the current segment of suitable frames. A \emph{segment} here is a contiguous set of frames, with no scene cuts. We add the longest good segments to the \emph{training} set for the respective identity until 5000 to 6000 frames are reached. All good segments beyond that make up the disjoint \emph{synthesis} set for this identity.

The \emph{training} set is processed with a monocular face reconstruction approach \cite{garrido2016reconstruction}, which encodes the facial performance as a sequence of parameter vectors.
 The vectors are then rendered to obtain the conditioning input that DVP learns to turn into RGB output again (\cref{fig:inference}). Thus for each identity, we obtain one DVP model that can render facial performances at photorealistic quality.
The input to such a model can be any arbitrary facial performance, also given as a sequence of parameter vectors. But to reliably avoid strong artefacts, we need to give facial performances as input that are close to the distribution that DVP saw during training (without, of course, using any of the training data!).
We simulate a faker that is able to synthesize such parameter sequences, by applying monocular reconstruction to the \emph{synthesis} set as well, thereby obtaining parameters that have the necessary properties. This is why our fakes mostly avoid clearly visible glitches, but still preserve the less noticeable artefacts that every GAN-based synthesis method inevitably exhibits.

For further details, including our modifications to DVP, we refer to our supplemental document. More synthesis results are shown in our submission video.

\section{Detecting High-quality Face Manipulations}
\label{sec:main}

\begin{figure}
\centering

\begin{overpic}[width=\linewidth]{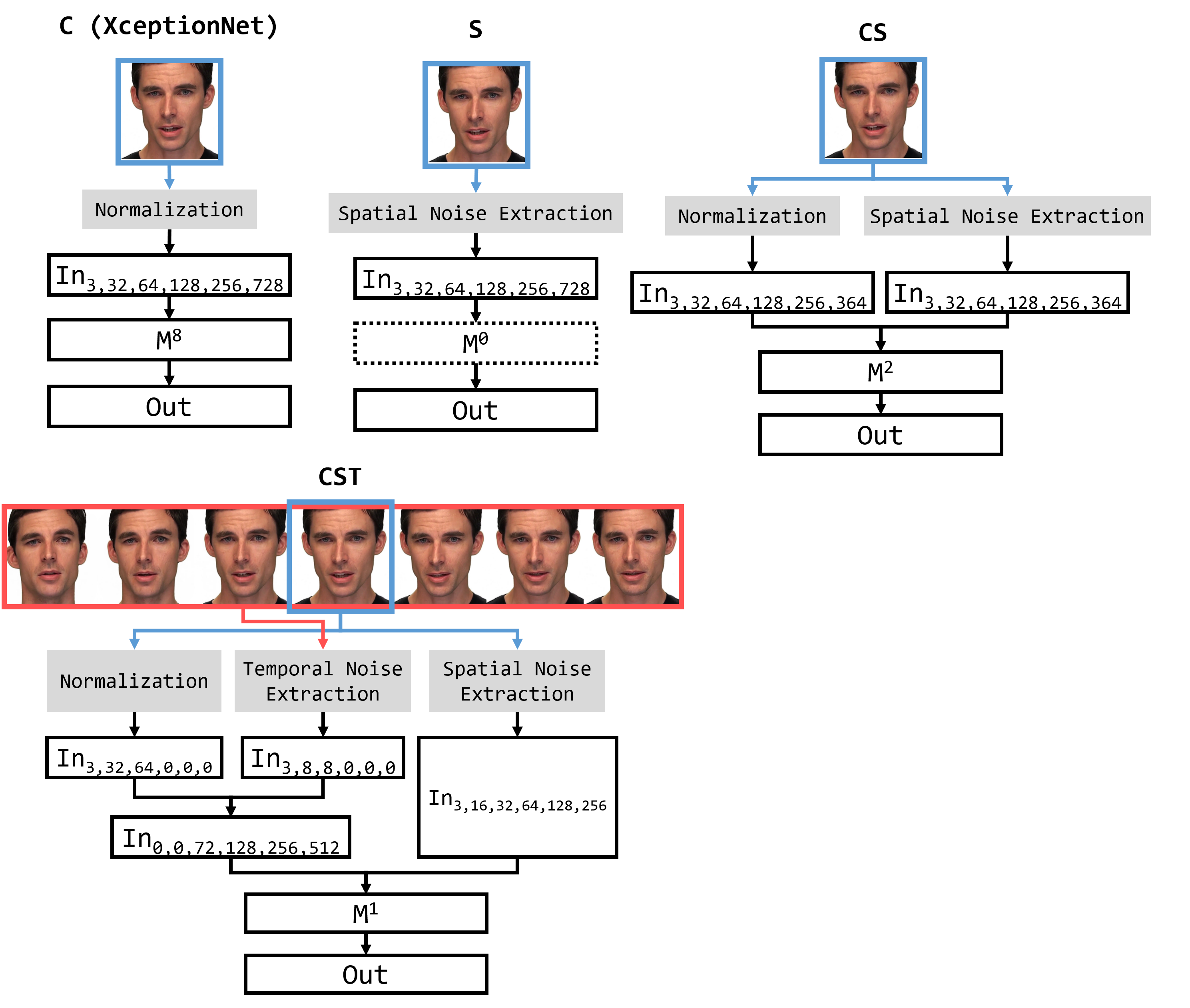}
     \put(60 , 5){ \includegraphics[width=0.35\linewidth]{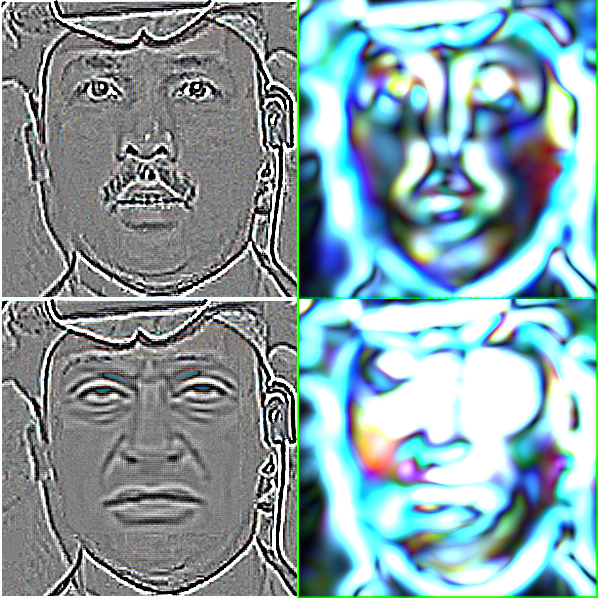}}  
	 \tiny     
     \put(64, 41){ Spatial noise}
     \put(81, 41){ Temporal noise }
     \put(58.5, 30){ \rotatebox{90}{Real} }
     \put(58.5, 13){ \rotatebox{90}{Fake} }
    
 \end{overpic}

	\caption{Our detectors extend XceptionNet~\cite{Chollet17} to a multi-stream classifier for combinations of color, spatial noise and even temporal features. 
	}
	\label{fig:detectors}
\end{figure}
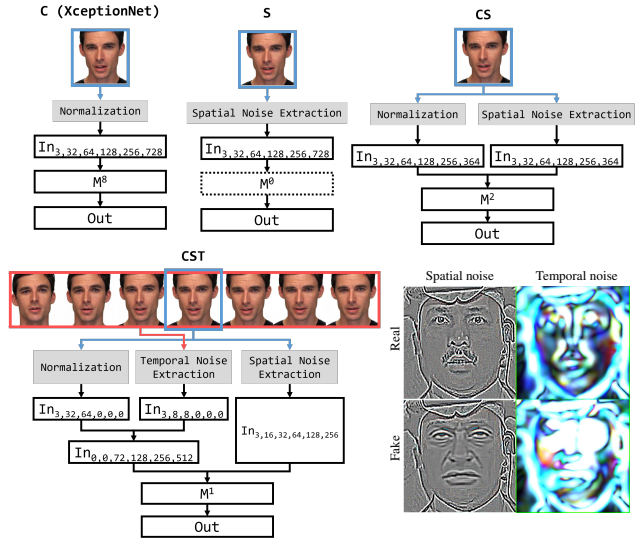

We consider XceptionNet \cite{Chollet17} a representative of existing face video forgery detection methods, because it ranked highest in FaceForensics++ \cite{roessler2019faceforensics++}. 
If \refhm{} is true, XceptionNet should perform worse on \DSName{} than it does on FaceForensics++. This expectation is justified because XceptionNet is a generic image classifier that  has not been designed for fake detection and thus 
should look for clearly visible artefacts in the image space.
 Since it nevertheless outperformed all other, detection-specific methods in \cite{roessler2019faceforensics++}, we want to enhance its ability to detect seemingly flawless fakes, without compromising its ability to detect strong artefacts.
We thus present a novel family of detectors that examine \emph{combinations} of multiple cues (\cref{fig:detectors}): the original RGB values, low-level spatial noise,  and temporal correlations.

XceptionNet consists of an entry flow $\In_{d\alpha\beta\gamma\delta\epsilon}$, a middle flow $\M$, and an exit flow $\Out$. Parameters $\alpha$, $\beta$, $\gamma$, $\delta$, and $\epsilon$ specify the number of features per convolutional layer (see supplemental), while $d$ specifies the number of input channels. One can denote XceptionNet as the function $C := \In_{3, 32, 64, 128, 256, 728} \circ \M^8 \circ \Out$,  applied to color images, yielding a score for class \enquote{real} and one for \enquote{fake}. 
 In the following, leading or trailing zeros in the indices of $\In_{d, \alpha, \beta, \gamma, \delta, \epsilon}$ disable the respective layers.

Repetitions of $\M$ drive up memory consumption and training overhead. To test whether 8 repetitions are actually necessary for forgery detection, we remove $M$ entirely:
\[\B := \In_{3, 32, 64, 128, 256, 728} \circ \M^0 \circ \Out\]
Since \DSName{} contains very few strong visual artefacts that $\C$ or $\B$ could easily pick up, we define $S$ to not classify frames $F \in \mathbb{R}^{299\times 299\times 3}$ themselves, but their spatially high-pass-filtered versions $\frac{1}{2} \cdot (F - g * F) + \frac{1}{2}$, where 
$g$ is a Gaussian kernel of size $5$, with standard deviation $\sigma = 1.1$. The architecture of $S$ is that of $B$.

Our combination of $C$ and $S$, \[\CS := (\In_{3, 32, 64, 128, 256, 364}, \In_{3, 32, 64, 128, 256, 364}) \circ \M^2 \circ \Out\] receives the same inputs as $\C$ and $\S$ and fuses the color and noise features just before entering $\M^2$, where the combined receptive field of the convolutional kernels has size $17 \times 17$ (see \cref{fig:detectors}).
We extend $CS$ to
\begin{align*}
&\PP_{\ct} \hspace{-0.2cm}&:= (\In_{3, 32, 64, 0, 0, 0}, \In_{3, 8, 8, 0, 0, 0}) \circ &\In_{0, 0, 72, 128, 256, 512} &\\
&\PP_{\cts} \hspace{-0.2cm} &\pushleft{:= (\PP_{\ct}, \In_{3, 16, 32, 64, 128, 256})}&\\
&\CST \hspace{-0.2cm}&  \pushleft{ :=\PP_{\cts} \circ \M^1 \circ \Out} &
\end{align*}
 which receives temporal features as a third input (\cref{fig:detectors}).
They are extracted as follows:
\begin{enumerate}
    \item Spatial Gaussian kernel (size $49$, $\sigma=7.7$), suppressing high spatial frequencies that motion would turn into temporal ones (e.g.  an edge sweeping over a pixel).
    \item Pixel-wise temporal high-pass filtering of the form $A_i := -\frac{1}{4} F_{i-1} + \frac{1}{2} F_i + -\frac{1}{4} F_{i + 1}$.
    \item Batch normalization.
    \item Amplitudes smaller than $t$ are dampened by computing
    $A'_i := \thr_t(A_i) - \thr_t(-A_i)$,
where the function    
    $\thr_t(A) := \frac{t}{10} \cdot (\ln{(1 + \exp(x))} + 
10 \cdot \sigmoid(x) )$ for  $x := \frac{10}{t} \cdot (A - t)$  is smooth and differentiable in $t$.
    \item Computation of temporal gradients: $G_i := |A_i - A_{i-1}|$.
    \item Temporal lowpass filtering (kernel $(\frac{1}{32}, \frac{1}{8},\frac{3}{16}, \frac{1}{8},\frac{1}{32})$).
\end{enumerate}
This process emphasizes unnaturally fast motions (i.e. quick changes from one frame to the next), often observed in forgeries. 
With the exception of step 1, all operations are pixel-wise. Step 2 suppresses low temporal frequencies, which are likely of natural origin. Steps 3 and 4 ensure that among the high frequency spikes we focus on those of a certain minimum amplitude, which are most likely artificial. Step 5 turns  oscillations between + and - into large positive values. Step 6 stabilizes the resulting signals, such that more output frames exhibit bright regions that the classifier can detect. For more information on $\thr_t$ see our supplemental material.

We are not aware of any existing face video forgery detectors that use such an approach.
We deliberately resist any reflex to make \enquote{everything trainable} in our %spatial and temporal
 feature extraction, to prevent it from overfitting to the training data. Only threshold $t$ is trainable, and our evaluation (\cref{sec:results:generalization}) shows that our detectors generalize better than completely trainable ones. 

The number of repetitions of $\M$ and the points at which we fuse streams were empirically chosen to maximize detection accuracy while not exceeding the 11GB of GPU memory in an NVIDIA 1080Ti. The resulting tradeoffs can lead to $\S$ performing slightly better than $\CS$ and $\CST$ on videos in which color and temporal information do not give a benefit over spatial noise, because the latter two cannot dedicate as much memory to spatial noise as $\S$ (\cref{sec:results}). On the other hand, spatial noise cues alone do not generalize as well as combinations with other types of information (\cref{sec:results:generalization}).

\section{Results}
\label{sec:results}

\newlength\Colsep
\setlength\Colsep{10pt}

Based on our new dataset \DSName{} and our novel family of detectors we can now investigate \refhm{}:

\myparagraph{State of the Art Techniques} We compare our detectors to a number of related techniques: Detector $C$, published as \enquote{XceptionNet} \cite{Chollet17}, performs best in the FaceForensics++ benchmark \cite{roessler2019faceforensics++}.  We evaluate $\mesonet$ \cite{Afchar2018}, $\misl$ \cite{Bayar18} and $\simplelinear$ \cite{durall2019unmasking} as they show good results in analysing low-level features. We also compare to $\easyspot$ \cite{wang2019cnngenerated}, a recent classifier that generalizes to unseen rendering methods.

\myparagraph{Preprocessings and training} Videos were preprocessed with one common pipeline before being fed into all detectors (temporally smooth face crops, rescaling to appropriate resolutions). Imbalances in the datasets were accounted for by weighted sampling. More details in supplementary material.

\subsection{Detecting highly photorealistic manipulations}
\label{sec:results:vanillatraining}

\begin{table}[t]
\caption{
Test accuracies and validation accuracy maxima of our detectors ($\B$, $\S$, $\CS$, $\CST$) and previous approaches.
}
\label{tab:vanillatraining}

\center
\begin{tabular}{|c|c|c|c|c|c|c|c|c|c|}
\hline               

\multirow{2}{*}{\arch} & \multirow{2}{*}{FF++}   &   \multicolumn{1}{|c|}{Deeper} &    \multirow{2}{*}{VFHQ}   \\

               &         &  Forensics  &                \\
\hline                                                     
     $\C$      & 99.23\% &  \textbf{98.64\%}    & 88.59\%               \\
     $\B$      & 99.34\% &  98.09\%    & 91.95\%               \\
     $\S$      & \textbf{99.38\%} &  98.21\%    & \textbf{99.45\%}               \\
    $\CS$      & 99.25\% &  98.32\%    & 97.12\%               \\
    $\CST$     & 99.35\% &  98.34\%    & 97.78\%               \\
   \mesonet    & 92.44\% &  97.25\%    & 76.73\%               \\
  \easyspot    & 75.55\% &  61.05\%    & 56.44\%               \\
\simplelinear  & 56.69\% &  62.88\%    & 61.98\%               \\
    \misl      & 95.82\% &  97.89\%    & 74.65\%               \\
\hline
\end{tabular}

\medskip
\note{Numbers averaged over 2 training runs.
}
\end{table}

To test \refhm{}, we trained all detectors on FaceForensics++ \cite{roessler2019faceforensics++} and DeeperForensics 1.0 \cite{jiang2020}, which both contain strong visual artefacts (\cref{fig:datasets}), as well as on \DSName{}, which does not (\cref{fig:images}).
\cref{tab:vanillatraining} confirms \refhm{}: With the exception of our novel detectors all accuracies are considerably lower for \DSName{}, than for previous datasets. In fact, XceptionNet ($\C$), the best detector in \cite{roessler2019faceforensics++} drops by more than 10\% and is even outperformed by $\B$, which is a reduced version of $\C$. Our detectors $\S$, $\CS$ and $\CST$ on the other hand  perform well on all three datasets. We observe that $\CS$ and $\CST$ perform not quite as well as $\S$. This is because they had to sacrifice some of the GPU memory that $\S$ can dedicate to spatial noise, in order to handle color and temporal features ( \cref{sec:main}). Since \DSName{} does not contain strong visual or temporal artefacts, this sacrifice does not pay off.

We remark that only our dataset is able to differentiate the best detectors from one another, while on previous datasets many detectors achieve close to 100\% accuracy.

\subsection{Generalization across manipulation techniques}
\label{sec:results:generalization}

Since the synthesis method for a forgery is often unknown, detectors should generalize to unseen methods.

To evaluate this ability, we train detectors on the \setff{} subsets $\setfs{} \cup \setnt{}$ (FaceSwap + Neural Textures \cite{thies2019}) and $\setftf \cup \setdf$ (Face2Face \cite{thies2016face} + Deep Fakes) and then test them on the subset they were \emph{not} trained on. 
We do not use \DSName{} for this experiment, because it contains only one manipulation technique and differs from other datasets in more respects than just the synthesis method (much higher resolution faces, no visible cues, etc). \setff{} is better suited here because all other factors can be kept constant when switching between synthesis methods.

\cref{tab:generalization} shows that training our detectors on $\setfs{} \cup \setnt{}$ makes them generalize well to $\setftf \cup \setdf$, where they outperform existing methods.
$\CS$ ranking higher than $\S$ and $\C$ suggests that \emph{combining} color and spatial noise can help generalization. 
The inverse, (train on $\setftf \cup \setdf$, test on $\setfs{}$), gives low accuracies for all detectors, suggesting that $\setfs{}$ contains artefacts not seen in $\setftf \cup \setdf$. The subset $\setnt{}$ seems to be easier to generalize to, with our detectors tending to outperform existing techniques and $\CST$ ranking highest.

Although detectors like $\C$ or \misl{} could theoretically learn the spatial filtering we hardcoded for $\S$, we see them perform considerably worse than $\S$ in \cref{tab:vanillatraining,tab:generalization}.

\subsection{Further results}
\label{sec:results:further}

Since no single dataset can cover all variations of forged video content (synthesis methods, image qualities, lighting conditions, camera angles, etc.), a robust detector should support training on a \emph{union} of datasets. We evaluated detectors on such a union and found ours to outperform previous ones, with $\CST$ ranking highest.
We also investigated the impact of the number of training identities and found our detectors to require fewer identities than previous methods. 
Detailed results can be found in our supplemental document.

\begin{table}[t]
\caption{Training on \setff{} subsets $\setfs{} \cup \setnt{}$ and $\setftf \cup \setdf$, with test accuracies for the opposite subsets.
}
\label{tab:generalization}

\center
\begin{tabular}{|c|c|c|c|c|}
\hline               
\multirow{2}{*}{\arch}  &   \multicolumn{2}{|c|}{Train: $\setfs{} \cup \setnt{}$} & \multicolumn{2}{|c|}{Train: $\setftf{} \cup \setdf{}$} \\
              &  \acc{\setftf} & \acc{\setdf} & \acc{\setnt} & \acc{\setfs} \\
\hline                                       
     $\C$     &    82.25\%          &   92.91\%           &  58.40\%             &   50.09\%     \\
     $\B$     &    90.40\%          &   95.15\%           &  66.03\%             &   50.27\%     \\
     $\S$     &    98.46\%          &   94.93\%           &  85.21\%             &   55.19\%     \\
    $\CS$     &    \textbf{99.46\%} &   98.74\%           &  86.74\%             &   51.78\%     \\
    $\CST$    &    98.91\%          &   \textbf{99.03\%}  &  \textbf{90.65\%}    &   56.77\%     \\
 %  $\CSnT$   &    99.21\%          &   99.18\%           &  85.42\%             &   57.65\%     \\
   \mesonet   &    89.65\%          &   71.94\%           &  73.10\%             &   49.76\%     \\
  \easyspot   &    77.91\%          &   80.03\%           &  84.40\%             &   \textbf{58.89\%}     \\
\simplelinear &    55.87\%          &   55.47\%           &  53.68\%             &   54.19\%     \\
    \misl     &    64.22\%          &   94.53\%           &  64.04\%             &   50.02\%     \\
\hline
\end{tabular}

\medskip
\note{Numbers averaged over 3 training runs. 
}
\end{table}

\section{Conclusion}
\label{sec:discussion}
In this paper, we have introduced \DSName{}, the first benchmark set for face video detection that provides a large number of manipulations a human would not be able to spot.
 Only with this dataset were we able to investigate whether current approaches to face video forgery detection are ready for the advent of synthesis methods that produce seemingly \enquote{perfect} results, confirming hypothesis \refhm{}.
To compensate for the shortcomings of existing detection approaches in this scenario, we have introduced a novel family of detectors  that combine spatial and temporal information in a way that has not been used in the area of face video forgery detection before. We have shown our detectors to outperform related methods both on previous datasets and on \DSName{}.

 While at first sight one might mistake the \enquote{intra-person} expression transfers in our dataset as harmless, recent works \cite{Suwajanakorn2017,Fried2019} demonstrate that even slight manipulations of this kind can have dramatic consequences. \DSName{} is the first and so far only benchmark dataset that allows their study. The absence of human-detectable artefacts in \DSName{} has the advantage of preventing detectors from learning to rely on their presence. This suggests that \DSName{} should be included in any \enquote{serious} detector training set and allows the detection community to prepare for future advances in forgery approaches already today.

\medskip
\myparagraph{Acknowledgement:}
\label{sec:acknowledgements}
This work was supported by the ERC Consolidator Grant 4DReply (770784).
% -------------------------------------------------------------------------
\bibliographystyle{IEEEbib}
\bibliography{refs}

\end{document}

\topmargin=0mm\sloppy

\title{\uppercase{VideoForensicsHQ: Detecting High-quality Manipulated Face Videos \\ -- Supplementary Material --}}
%
% ---------------
\name{ \parbox{\linewidth}{\centering Gereon Fox, Wentao Liu, Hyeongwoo Kim, Hans-Peter Seidel, \\ Mohamed Elgharib \& Christian Theobalt}}
\address{Max Planck Institute for Informatics, Saarland Informatics Campus \\ \small\texttt{\{gfox,wliu,hyeongwoo.kim,hpseidel,elgharib,theobalt\}@mpi-inf.mpg.de}}
\maketitle

\section{Synthesis details}

\begin{figure*}
        \centering
         \includegraphics[width=0.85\linewidth]{pics/mosaic_fullpage}
         \caption{Forgeries from \DSName. 
        }
        \label{fig:vfhq}
\end{figure*}

We mined the authentic raw footage for our dataset from \cite{Kim19NeuralDubbing} (\setSA{}), RAVDESS \cite{Livingstone2015TheRA} (\setRD{}) and YouTube (\setYT{}). \cref{tab:stats} lists the sizes of these subsets.

The original version of DVP \cite{kim2018DeepVideo} cannot handle dynamic backgrounds and works at a fixed resolution of $256^2$. We thus prepare training frames by cropping them around the face and masking out the background (see pipeline figure in the main paper), in order to make the network focus all its capacity on the face region. The resulting square images are scaled to resolution $256^2$.
Instead of a separate conditioning image for the eye-gaze (like in \cite{kim2018DeepVideo}) we overlay the eye gaze rendering on the face rendering.
 We use temporal supervision for DVP, by means of \enquote{temporal windows}, as proposed in \cite{kim2018DeepVideo}: The discriminator sees temporal volumes of 5 frames for most of \setRD{}. We found the temporal window to not improve quality considerably, which is why \setYT{} and \setSA{} were synthesized with window size 1.
We train our DVP models for up to 200 epochs, estimating the mean squared photometric error against ground truth on the validation set. The model with the smallest error is used for synthesis.
Since the facial performances we are rendering at synthesis time have been reconstructed from real footage, we know the coordinates of the face region in that footage. This allows us to alpha-blend the DVP output into those original frames.

\cref{fig:vfhq} shows more forgery examples from our dataset.

\begin{table}[t]
\caption{\DSName{} subsets}
\label{tab:stats}
\center
\begin{tabular}{|c|r|r|}
\hline
Subset & Fake frames & Real frames \\
\hline
group\#1 & 60,058 & 119,992  \\
group\#2 & 74,765 & 190,259  \\
group\#3 & 192,150 & 1,029,592  \\
total    & 32,6973 & 1,339,843 \\
\hline
\end{tabular}

\end{table}

\section{User Study}
\label{sec:user}

We conducted a user study to asses the quality of our fakes compared to FaceForensics++ \cite{roessler2019faceforensics++}.
We randomly selected 13 manipulated videos from \DSName{} and 13 manipulated videos from the \enquote{Neural Textures} subset of  FaceForensics++ \cite{roessler2019faceforensics++}, created with the reenactment technique by Thies~\etal~\cite{thies2019}. Other approaches in FaceForensics++ produce fakes with much more visible artefacts (see Figure 2 %\cref{fig:datasets}
 in the main manuscript). In addition, we randomly selected 6 unmodified videos from \DSName{} and 7 from FaceForensics++.

\begin{table}[t]
\caption{User study results}
\label{tab:userstudy}
\center
\begin{tabular}{|c|r|r|}
\hline
 \multirow{2}{1cm}{Source}    & \multirow{2}{1cm}{Rated \enquote{fake} }  & \multirow{2}{1cm}{Rated \enquote{real} }   \\
 & &  \\
\hline
Real videos & 15.0\% & 85.0\%\\

\enquote{Neural Textures} fakes \cite{roessler2019faceforensics++} & 85.7\% & 14.3\%\\
\DSName{} fakes & 34.2\% & 65.8\%\\
\hline
\end{tabular}
\end{table}

In total our study contains 39 videos, randomly shuffled for each participant. For each video, we recorded the answer to the question \enquote{Does the video look real or fake?}. 
Most participants were computer scientists, with little-to-no knowledge of face manipulation techniques. 61 subjects participated in the study. On average, fakes from \DSName~were rated real $65.8\%$ of the time, and fakes from FaceForensics++ were rated real only $14.3\%$ of the time. \cref{tab:userstudy} lists the full results. We note that unmodified videos were also rated as manipulated $15\%$ of the time, which reflects a baseline error level in human detection performance.
We also asked participants what made them flag a video as modified.
Some of the most common responses were:
\begin{enumerate}
\item Various visual artefacts, especially in mouth interior 
\item Non-natural eye movement 
\item Body movements or hand gestures  not matching speech 
\item Non-natural mouth-related movements e.g. lips being tight when they should not be, deforming/dislodging jaw, etc.
\item Incorrect audio-lip synchronization 
\item A single glitch occurring over 2-3 seconds
\item Spoken language not matching language of written text
\end{enumerate}

\section{Detection details}

\subsection{Parametrization of XceptionNet}

\begin{figure*}
    \centering
    \includegraphics[width=0.8\linewidth]{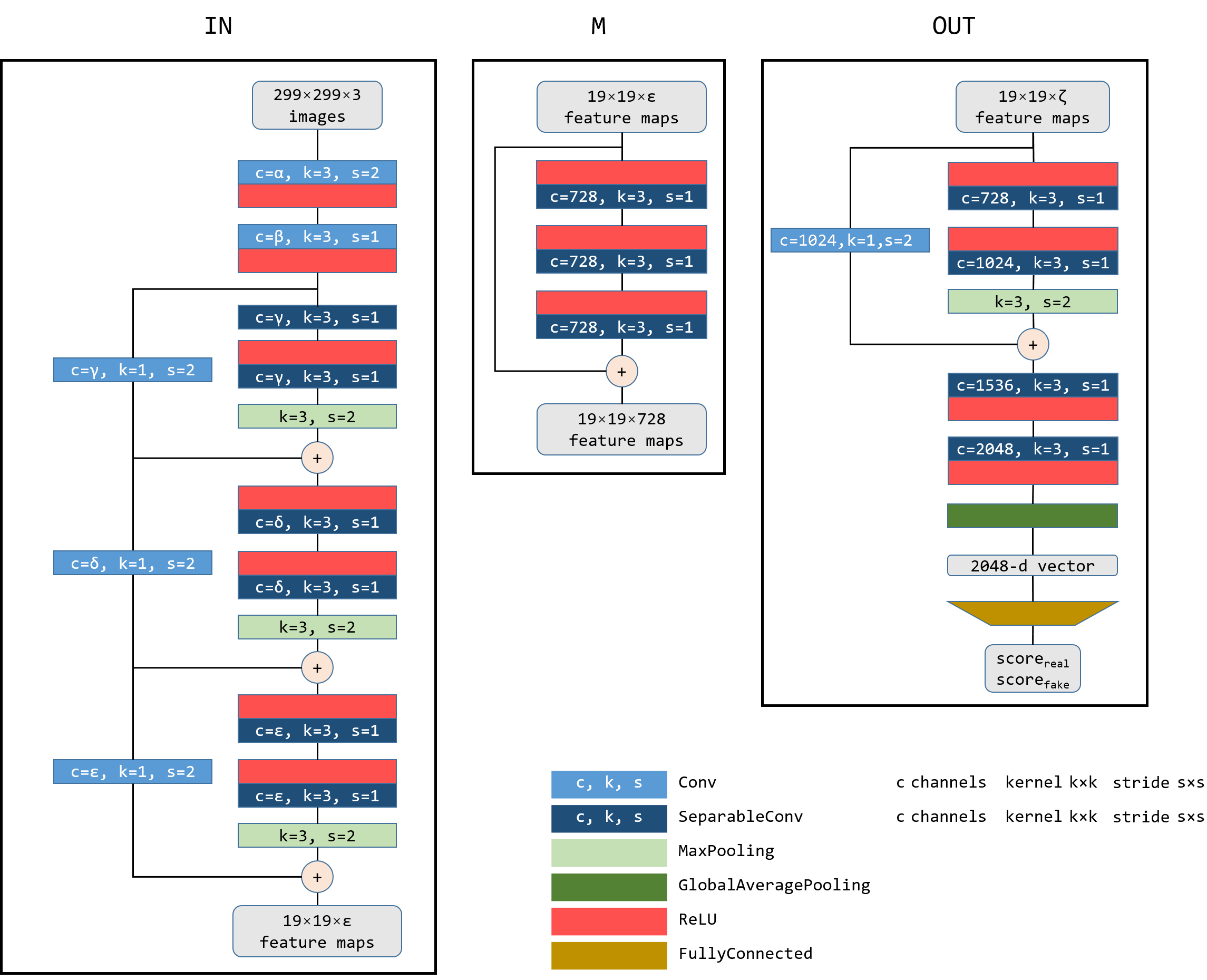}
    \caption{The building blocks of XceptionNet \cite{Chollet17} are the basis of our multi-stream detectors (see Figure 4 of the main manuscript). In order to trade memory capacity between multiple streams and fuse them at the right locations, we changed the numbers of features in each layer of each instance of such a building block. $\epsilon$ in $\M$ and $\zeta$ in $\Out$ are determined by the number of output feature in the preceding block.}
    \label{fig:nnblocks}
\end{figure*}

In order to build our detectors, we have generalized XceptionNet \cite{Chollet17}, by parametrizing various dimensions of its architecture (see main paper). \cref{fig:nnblocks} clarifies the meaning of each parameter we introduced.

\subsection{Temporal filtering}

\begin{figure}[t]
\center
\includegraphics[width=\linewidth, trim=30 30 40 50, clip]{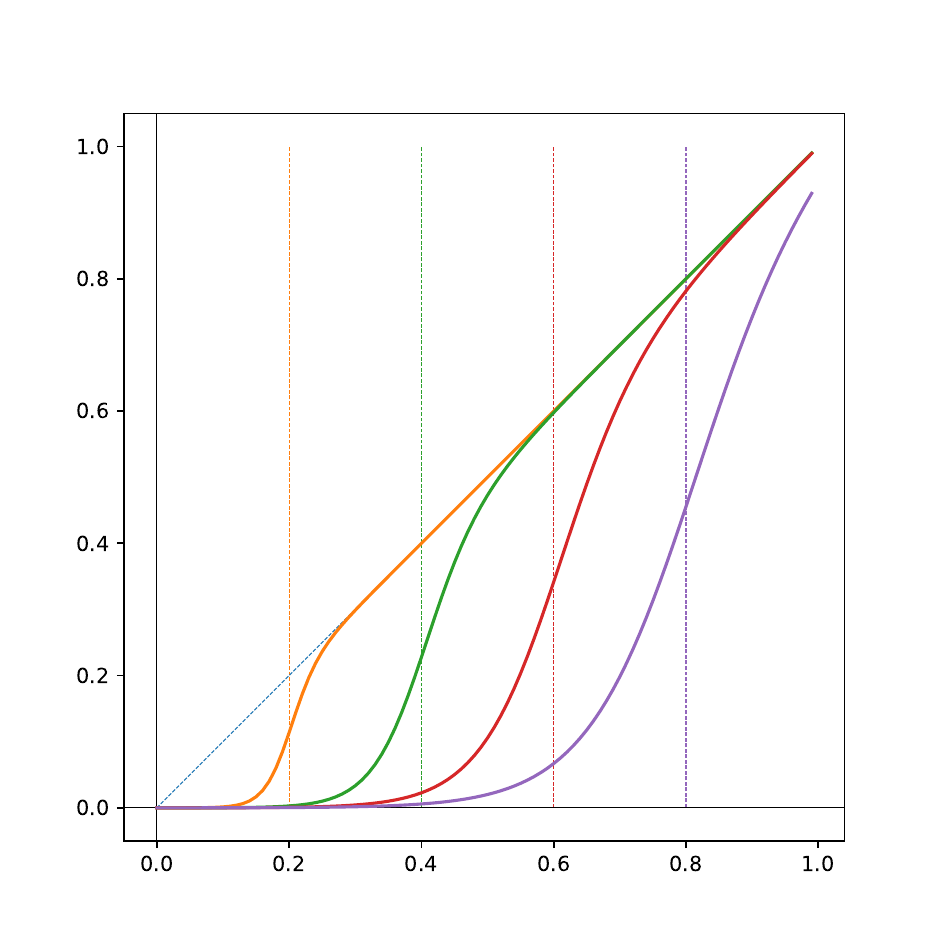}
\vspace{-0.5cm}
\caption{Graphs of $\thr_t$, for $t \in \{0.2, 0.4, 0.6, 0.8\}$.}
\label{fig:thr}
\end{figure}

The function $\thr_t$ is supposed to dampen low amplitudes of a signal. The parameter $t$ is a threshold specifying which amplitudes are to be dampened. \cref{fig:thr} plots $\thr_t$ for different values of $t$. The function is differentiable in $t$, such that the training process can automatically determine a good position and shape for the \enquote{cliff}. 

\subsection{Detection preprocessing}

All training and test data for all detectors is preprocessed by one common pipeline: Face bounding boxes are computed using \emph{dlib} \cite{dlib09},
with temporal smoothing of their coordinates. We extract constant-size square bounding boxes, scaled to resolution $299^2$. We resample videos at 25fps. Frames for which no face bounding box can be found are omitted. For \mesonet{} we scale the resulting frames to $256^2$, whereas for \simplelinear{} we compute a $209$-dimensional feature vector as specified in \cite{durall2019unmasking}.

\subsection{Detection training}

The Xception-based detectors $\C$, $\B$, $\S$, $\CS$ and $\CST$ are all trained with stochastic gradient descent (momentum $0.9$, weight decay $10^{-5}$), multiplying the initial learning rate of $0.03$ with factor $0.97^{0.1}$ per epoch. For $\CST$ we initialize $t$ with  $\frac{1}{40}$.
Previous detectors are trained as specified in their publications, except for \easyspot{} and \simplelinear{}:
\easyspot{} is claimed to generalize well to unseen rendering methods. We thus use its pretrained weights and merely optimize a threshold on its singular output value, based on the ROC curve over the samples that were seen within one epoch of training. We perform this optimization for 5 epochs and average the 5 resulting thresholds.
For every training batch of $\simplelinear$ \cite{durall2019unmasking}
we optimize a new SVM model on the Fourier features. At validation and test time we average the predictions of all SVM models obtained in this way.

All detectors are trained with batch size $24$, except for \mesonet~ ($512$) , \simplelinear~($512$) and \misl~($256$). Except for \easyspot, all methods are trained with a hard limit of 100 epochs. We stop training earlier if
5 epochs with a validation accuracy of more than 99\% have been seen (not necessarily consecutively). The model with maximal validation accuracy is used at test time.

To account for imbalances in the datasets, we randomly sample 10\% of the training frames and 20\% of the validation frames in every epoch. Sampling here means to first uniformly select a class (\enquote{real}/\enquote{fake}), then a subset (which is relevant for \DSName{} because it consists of three different groups), then a subject and then one of the sequences for this subject. Frames are sampled uniformly from sequences. Since \simplelinear{} is not designed for the amounts of data resulting from the aforementioned sampling rates we lower them to 0.5\% training and 1\% validation samples for this method.

This training procedure is the reason why in Table 1 of the main paper, the accuracies we report for \mesonet{} and \misl{} on \setff{} are slightly lower than the ones reported in \cite{roessler2019faceforensics++}. However, the detectors based on XceptionNet~\cite{Chollet17} ($\C$, $\B$, $\S$, $\CS$, $\CST$), do not seem to be impacted by this, which we interpret as a strength of XceptionNet-based architectures.

At test time, we evaluate \emph{all} frames of the test set in which a face could be found, but weigh per-frame predictions by the probability of a frame  being sampled according to above sampling process.

\section{Further results}

\subsection{Training on a union of datasets}

Since no single dataset can cover all variations of forged video content (synthesis methods, image qualities, lighting conditions, camera angles, etc.), a robust detector should support training on a \emph{union} of datasets.
To evaluate how well detectors handle  this setup we have trained them on the union of \setff{}, \DSName{} and \setpdfdc{}~\cite{dolhansky2019} and tested them on \setff{} and \DSName{} (\cref{tab:diverse}\footnote{\cite{dolhansky2019} contains very challenging perspectives and lighting conditions, as well as fast motion, making our preprocessing struggle to the point that it becomes the limiting factor of accuracy: All detectors, ours and previous, achieve about 80\% test accuracy.}). Compared to training on only one single dataset (see main paper), the  task is now hard enough to also differentiate \emph{our} detectors from one another: $\B$ again performs better than $\C$. $\S$ and $\CS$ are on par. $\CST$ can now demonstrate the benefit of temporal information, ranking highest on both test sets.

\begin{table}[t]
\caption{Test accuracies after training on the union of \setff{}, \DSName{} and \setpdfdc{}~\cite{dolhansky2019}.
}
\label{tab:diverse}
\center
\begin{tabular}{|c|c|c|}
\hline
\multirow{2}{*}{\arch} & \multicolumn{2}{|c|}{Test accuracy}     \\
        & \setff & \DSName \\
\hline
     $\C$     &   95.90\%    &    78.99\%     \\
     $\B$     &   96.16\%    &    80.02\%     \\
     $\S$     &   97.69\%    &    88.94\%    \\
    $\CS$     &   98.01\%    &    87.91\%    \\
    $\CST$    &   \textbf{98.67\%}    &    \textbf{90.63\%}    \\
   \mesonet   &   74.34\%    &    76.65\%    \\
  \easyspot   &   75.88\%    &    56.75\%    \\
\simplelinear &   54.20\%    &    55.51\%    \\
    \misl     &   90.02\%    &    78.85\%    \\
\hline
\end{tabular}
\medskip
\end{table}

\subsection{Impact of training corpus size}
\label{sec:results:corpussize}

\begin{figure}[t]
\center
\includegraphics[width=\linewidth]{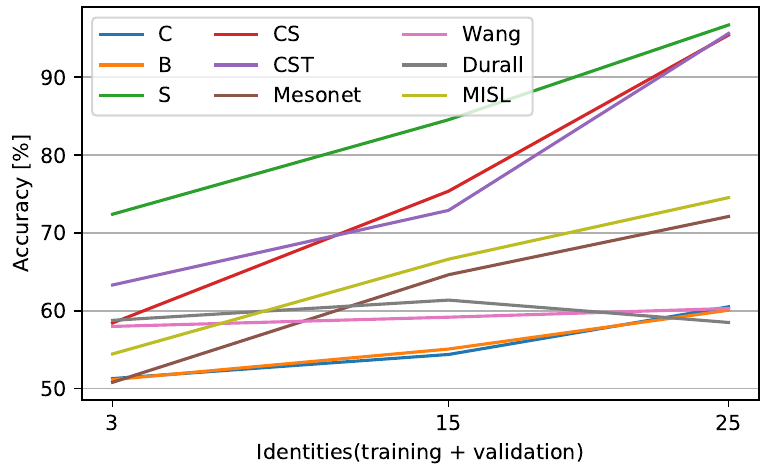}
\caption{Average test accuracies achieved by models that were trained on  \DSName{} subsets containing different numbers of identities, see \cref{sec:results:corpussize}. 
}
\label{fig:fewshot}
\end{figure}

\DSName{} contains only 45 identities, while \setff{} contains 1000 identities. This raises the question of how many identities are necessary to train a detector.

We have thus randomly sampled small  training sets from
\DSName{}, with different numbers of identities.
Detectors were trained
on these subsets and then tested on random test sets of 15 identities each (disjoint from the training sets). \cref{fig:fewshot} shows the average accuracies after training on different numbers of identities. For each number of identities we have sampled 3 to 5 different training sets.

We observe that the best detectors achieve close to 100\% test accuracy already for training corpora of only 25 identities (training + validation), which is much fewer than the total number of identities in \DSName{}, providing evidence that our dataset is sufficient to generalize to unseen identities, and that our detectors do \emph{not} overfit to the training identities.

% -------------------------------------------------------------------------
\bibliographystyle{IEEEbib}
\bibliography{refs}